# Clustering by Hierarchical Nearest Neighbor Descent (H-NND)

Teng Qiu          Yongjie Li[*]
(liyj@uestc.edu.cn)
University of Electronic Science and Technology of China, Chengdu, China

**Abstract:** previously in 2014, we proposed the **Nearest Descent** (**ND**) method, capable of generating an efficient Graph, called the **in-tree** (**IT**). Due to some beautiful and effective features, this IT structure proves well suited for *data clustering*. Although there exist some *redundant edges* in IT, they usually have salient features and thus it is not hard to remove them.

Subsequently, in order to prevent the seemingly redundant edges from occurring, we proposed the **Nearest Neighbor Descent** (**NND**) by adding the "Neighborhood" constraint on ND. Consequently, clusters automatically emerged, without the additional requirement of removing the redundant edges. However, NND proved still not perfect, since it brought in a new yet worse problem, the "*over-partitioning*" problem.

Now, in this paper, we propose a method, called the **Hierarchical Nearest Neighbor Descent** (**H-NND**), which overcomes the over-partitioning problem of NND via using the *hierarchical strategy*. Specifically, H-NND uses ND to effectively merge the over-segmented sub-graphs or clusters that NND produces. Like ND, H-NND also generates the IT structure, in which the redundant edges once again appear. This seemingly comes back to the situation that ND faces. However, compared with ND, the redundant edges in the IT structure generated by H-NND generally become more salient, thus being much easier and more reliable to be identified even by the simplest edge-removing method which takes the edge length as the only measure. In other words, the IT structure constructed by H-NND becomes more fitted for data clustering. We prove this on several clustering datasets of varying shapes, dimensions and attributes. Besides, compared with ND, H-NND generally takes less computation time to construct the IT data structure for the input data.

## 1 Introduction

Previously in (*1*), we proposed a physically-inspired clustering method by mimicking the moving behavior of particle swarm. Imagine that **(i)** each data point behaves like a particle that can generate a field whose magnitude declines with the distance, and that **(ii)** the fields from different data points can be added up in the same places. As a result, this point (or particle) system would generate an "uneven" space[1], and in turn, this uneven space could enforce the point system to evolve. Intuitively, the points in this system tend to move along the *descending direction* of potential (i.e., from higher to lower potential areas), and eventually get clustered in the places of the locally lowest potentials. Accordingly, we devised the method as follows.

First, we *simplified* the *descending tendency* of points as such a general rule:

---
[1] "uneven" means the potentials varies in places. Usually, points in the dense areas would have lower potentials.

"descend" to the nearest point, or called **the nearest descent** (**ND**). Namely, among the nodes in the decreasing direction of potential, each point was designed to choose the nearest one to "descend" to. If we view each point *i* as a *node,* and link each node (except the node with the lowest potential) to *the node it descends to* (denoted as the **Parent Node** of node *i*) by a directed *edge*, then the ND rule can result in a *Graph* structure.

In Graph Theory, the Graph is composed of the nodes (referring to the data points) and the edges (defining the linkage relationships between nodes). In fact, the Graph constructed by ND is a special and fully connected graph, call the ***in-tree*** (***IT***) in Graph theory, which was proven in (*1*). IT itself has several salient features: **(i)** it is a very sparsely connected graph, in which the number of the directed edges is just one less than that of the data points in the dataset; **(ii)** there is wonderful *order* inside it, that is, every other node has *one and only one* directed path to reach the root node. Besides that, the IT structure constructed by ND method, as Fig. 1 shows, has two additional yet meaningful features: **(iii)** the underlying cluster structure in the datasets can be well captured by IT, except very few (usually at most one) *redundant edges* between clusters requiring to be removed; **(iv)** Those redundant edges usually have very salient features and thus it is not hard to identify them. All these features make the IT structure constructed by ND well suited for the task of data clustering, a *basic*, *classic* yet still *challenging* task in statistics, data mining and machine learning (*2*), aiming at discovering the underlying cluster structure in the datasets and assigning the data points belonging to the same clusters into the same groups.

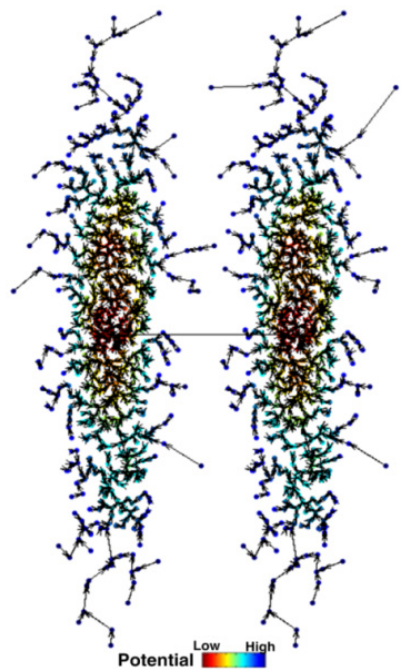

**Fig. 1 The IT structure of one data set.** The two elongated clusters in the dataset is well captured by the IT structure in a sparse and effective form, and there exists one redundant yet salient edge bridging over the two clusters. The colors on points denote the estimated potential values.

Based on the IT structured generated by ND, two steps are needed for clustering purpose: **(i)** remove the redundant edges in IT (this will divide IT into several unconnected sub-graphs); **(ii)** search the root nodes and assign the nodes belonging to the same root nodes into the same groups.

For the 1st step, since the foremost feature for the redundant edges is that they are usually relatively longer than other edges (see the redundant edge between the two clusters in Fig. 1), the simplest method to determine them is to rank the edge lengths in decreasing order and take the ones with relative large edge lengths as the redundant edges. This works well at least for the dataset in Fig. 1. Beside this simple method, we have in fact devised a bunch of more reliable methods to determine the redundant edges, either interactively or automatically, such as ***IT-map*** (3), ***IT-Dendrogram*** (4), ***G-AP*** (5) and other methods in (*1*). Once removing the redundant edges, we can immediately obtain several unconnected sub-graphs, and each sub-graph (still being an IT structure) represents one cluster.

For the 2nd step, the root searching process can be simply performed by searching along the directions of edges. As a result, each node will find the root node of its cluster. The nodes reaching the same root nodes (functioning like the cluster centers) are thus assigned into the same groups. Unlike the iterative methods as K-means (6) and Mean-shift (7, 8), this searching process on the graph (*functioning as a map*) is almost negligible in computation time (*1*), since the searching paths are already specified in the graph and the searching process can be operated in parallel for all points.

## 2 Motivation and Idea

In this work, instead of proposing new method to determine the redundant edges, our ***motivation*** here is to modify the rule of ND so that the redundant edges become more salient and thus can be more easily and reliably determined even by the simplest edge-removing method (i.e., the method only referring to the edge length).

In fact, according to ND, the start node of each redundant edge is the *extreme point* (referring to the node of the locally lowest potential, or the root node of one sub-graph) in one cluster, while this is generally not true for the end node. The end node should have the lower potential than the start node, whereas not necessary to be extreme node in another cluster. In Fig. 2, we illustrate this more clearly. As Fig. 2A shows, we take a 1-dimensional dataset with two main clusters as an example. Its IT structure constructed by ND is shown in Fig. 2B. Note one difference: in Fig. 2B the values of the estimated potentials on all points are denoted by an additional axis (the vertical axis), while in Fig. 1A they are denoted by different colors. In Fig. 2B, we can see that data points from two clusters are respectively organized into two bowl-like shapes. And the redundant edge (in red) starts from the bottom node (or the extreme node) of the left bowl and ends in the right bowl, yet obviously the end node is not at the bottom. ***Specifically***, ***our motivation can be interpreted as***: to make the end node of the red edge also be the bottom node of the corresponding bowl, as Fig. 2D shows. Obviously, at least one good thing for this change is that, compared with Fig. 2B, the redundant edge in Fig. 2D becomes longer, with larger difference with the rest edges.

The ***difficulty*** for this motivation is that the new rule should only enhance the feature of the redundant edges while making no or insignificant change on the rest ones.

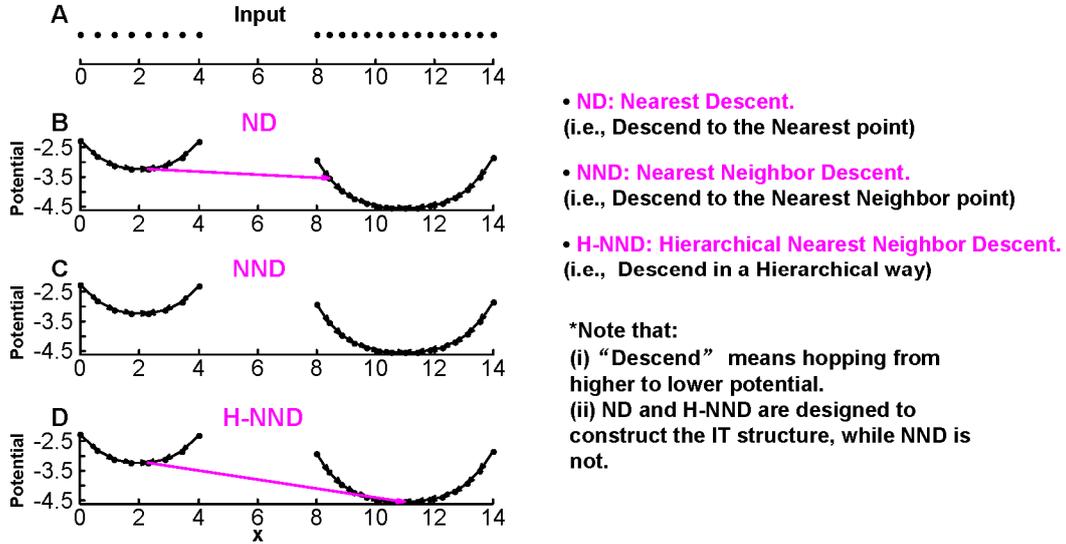

**Fig. 2 The illustration for the motivation.** (A) The input 1-dimensional data points. The graphs generated by ND, NND, H-NND are shown in (B), (C) and (D), respectively. The horizontal axis in (A~D) denotes the coordinates of the input data points. The vertical axis added in (B~D) denotes the potential estimated on each data point. The red edges in (B) and (D) denote the *redundant edges* that need to be removed.

The solution in this paper is in fact based on another method proposed by us in (9), in which we have already started to modify the ND rule while in a different purpose. The method there, denoted as the ***Nearest Neighbor Descent*** (***NND***), could prevent the redundant edges from occurring in the generated graph (Fig. 2C) via constraining the ND rule in the *Neighbor Graph* (or the graph space). Namely, compared with ND, the additional letter "N" in the middle of NND refers to the neighborhood constraint added on ND.

NND was intended to make the rule more consistent with the real physical circumstance, and thus it took into account the physical space in which no particle could travel through space to reach another place. In practice, as Fig. 3A shows, NND approximated the physical space by the neighbor graph, which severs as the neighborhood constraint while performing the ND rule. This leads to the result as shown in Fig. 3B.

Admittedly, one thing fascinating for NND is that, due to the neighborhood constraint, the redundant edges do not emerge, as Fig. 3B or Fig. 2C shows, and thus the effort of removing the redundant edges is not needed. Another less obvious benefit also brought by the neighborhood constraint is about computation complexity, that is, the searching of the parent nodes (i.e., to which each node should "descend") are narrowed in the local scopes rather than the whole dataset, which means a huge reduction in computation time while processing the datasets with large numbers of data points. Nevertheless, the neighborhood constraint could bring in a new yet nontrivial problem: the ***over-partitioning problem***. As revealed also in Fig. 3B, certain cluster is partitioned into more than one part. This is not surprising at all, since the neighborhood constraint would make the locally extreme nodes (the nodes with lower potentials than all their neighbors) directly become the roots of the generated sub-graphs (or the cluster centers of the generated clusters). Therefore, when the estimated potentials on nodes are of poor representation of the underlying density distribution, the number of the obtained extreme nodes will be more than that of the main clusters inside the dataset, and in this case the over-partitioning problem will occur.

However, NND sheds lights on the goal of this paper. On the one hand, the extreme nodes identified by NND bring the over-partitioning problem. On the other hand, they could also be a treasure, if we make use of it effectively. For instance, what if we rerun the ND rule on the extreme nodes obtained by NND? In other words, what if we let each extreme node continue to "descend" to its nearest point among the extreme nodes. Yes, a wholly linked graph (Fig. 3C) will be constructed. Obviously, this newly generated graph is still the IT structure and both the start and end node of the undesired edges (the red one in Fig. 3C) in it are sure to be the extreme nodes. Namely, our initial goal mentioned in Section 2 is achieved. Besides, it is also worth noting that the over-partitioned clusters in Fig. 3B are effectively linked together in Fig. 3C, which means the over-partitioning problem that NND faces is also largely solved by the proposed method.

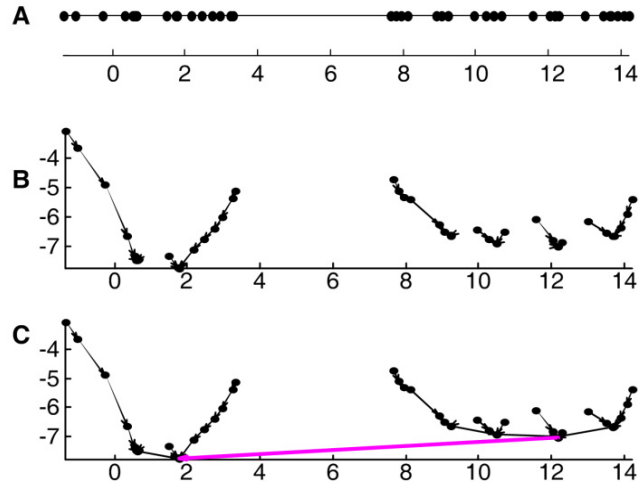

**Fig. 3 An illustration for the process of HNND.** (A) Construct the neighbor graph. (B) Run NND. (C) Run ND to merge the root nodes generated in (B). Red edge is the redundant edge.

## 3 The proposed method: H-NND

Therefore, the key feature for the proposed method is that of the *hierarchical strategy*, i.e., ND is performed in the second stage on the extreme nodes obtained from NND in the first stage. We denote the proposed method as the ***Hierarchical Nearest Neighbor Descent*** (***H-NND***), containing the following four steps:

**Step 1, construct the neighbor graph**[2].

The input data points $\Psi = \{1, 2, \cdots, N\}$ can be constructed into the K-Nearest-Neighbor (K-NN) graph, Delaunay Triangulation (DT) (10) or the minimal spanning tree (MST) (11). Here, we use $\eta^i$ to denote the neighbor nodes of each node $i$.

**Step 2, compute the potentials:** $P_i = -\sum_{j \in \Psi} e^{-d_{i,j}/\sigma}$, where $d_{i,j}$ denotes the pair-wise distance between nodes $i$ and $j$.

**Step 3, run NND.**

First, for each node $i \in \Psi$, identify the neighbors of lower potential: $J^i = \{j \mid P_j < P_i, j \in \eta^i\}$. $J^i$ is called the *candidate* parent nodes of node $i$, which can be null for some nodes (denoted as data set $E = \{i \mid J^i = \varnothing\}$). Then, for each node $i \in \Psi - E$, set the nearest neighbor from $J^i$ as its parent node**:**
$$I_i = \arg\min_{j \in J^i} d_{i,j}, \quad i \in \Psi - E.$$

**Step 4, run ND to merge the root nodes.**

For the remaining nodes $i \in E$, compute again their candidate parent nodes set while in the new rule $J^i = \{j \mid P_j < P_i, j \in E\}$. Only one node (with the minimum potential) in $E$ will have null $J^i$. Set its parent node as $I_i = i$. For each of the other nodes in $E$, set its parent node as:
$$I_i = \arg\min_{j \in J^i} d_{i,j}, \quad i \in E.$$

Note that, in practice, we need to consider the case when there are points in $J^i$ with the same potential or with the same distance to node $i$, therefore the true forms are slightly complex than that in step 4:
$$I_i = \min \arg\min_{j \in J_i} d_{i,j},$$

where $J^i = \{j \mid P_j < P_i, j \in E\} \cup \{j \mid P_j = P_i \,\&\&\, j < i, j \in E\}$. Based the neighborhood graph constructed or the input Graph-based datasets (or the complex networks), NND is first performed to obtain the result of the first layer of the hierarchy. This will result in a sparse yet not fully connected graph, usually with more than one unconnected sub-graphs. Then, in the second stage, ND is performed to merge those sub-graphs into a fully connected graph (i.e., the IT). Note that it is only needed to link the root nodes of those sub-graphs by ND. Therefore, the whole process can be viewed as a process of the Graph transformation (or the networks evolutions), resulting in a sparse (or economic) yet effective one at last.

The following steps for clustering purpose are the same as the two steps of the ND-based clustering process mentioned in Section 1, i.e., edge removing and root searching. By plotting the lengths (the distances between each point with its parent node in decreasing order) of all edges, denoted as the ***Edge Plot***, we can immediately find out the redundant edge underlying in the constructed IT structure. Note that, in the Edge plot, each edge length represent the distance between each data with its parent node in the original space. For example, in Fig. 1, the edge lengths in the Edge Plot refer to the lengths of the edges plotted in it, while in Fig. 3C (an additional vertical axis is added), the edge lengths in the Edge Plot refers to the lengths of the mappings of all the edges on the horizontal axis rather than the edges plotted. After removing the redundant edges, each sub-graph actually represents one cluster.

---

[2] Note that if the input datasets are already graph-based datasets, step 1 is not needed.

There are ***at most*** two parameters in H-NND: one is the parameter $K$ that may be involved in Step 1, if the input data is not the graph dataset and the data points in it are chosen to be constructed into the K-NN graph; the another is the parameter $\sigma$ involved in Step 2, if we use the kernel-based method to estimate the potentials on nodes (*1*). However, when we use the nonparametric graph (e.g., DT) in step 1, and the non-kernel-based way to estimate the potential in step 2, the number of the parameters can become zero. We will show this in Section 4.4.

## 4 Experiments

We tested the performance of H-NND by clustering a wide range of data sets[3] with varying shapes, dimensions and attributes (see Table 1). Two (the 8th and 10th) of them are collected from the real life; the others are synthetic datasets. Three (the 8th, 9th and 10th) are high-dimensional datasets; the others are 2-dimensional (2D). It is challenging (*2*) for one clustering method to deal with different kinds of datasets. For instance, the classical method like K-means (*6*) and the modern method like AP (*12*) are only able to detect spherical clusters, and thus it is hard for them to cluster most of the 2D datasets in Table 1. Besides, the Iris (the 8th) and Mushroom (the 10th) datasets are widely used to check the performance of the clustering methods (*13*). In the experiments mentioned below we achieved very excellent (almost perfect) results on all these datasets, and in all experiments we used the simplest edge removing method (i.e., via Edge Plot).

**Table 1. The Test Datasets**

|   | Dataset | Attribute | $d$ | $N$ |
|---|---|---|---|---|
| 1 | Aggregation | real number | 2 | 788 |
| 2 | Flame | real number | 2 | 240 |
| 3 | Spiral | real number | 2 | 600 |
| 4 | Flame | real number | 2 | 240 |
| 5 | Science-14 | real number | 2 | 4000 |
| 6 | S1 | real number | 2 | 5000 |
| 7 | S4 | real number | 2 | 5000 |
| 8 | **Iris** | real number | **4** | 150 |

---

[3] Downloaded from http://cs.joensuu.fi/sipu/datasets/; http://archive.ics.uci.edu/ml/; and http://people.sissa.it/~laio/Research/Res_clustering.php

| 9 | D32 | real number | **32** | 1024 |
| 10 | **Mushroom** | **Character** | **22** | 8124 |

*d*: dimensionality. *N*: number of data points.

**4.1 Tests on 2-dimensioal datasets**

    The results testing on the 2D synthetic datasets are shown in Fig. 4. Its left column shows the Edge Plots of the IT structures generated by the proposed H-NND method, and its right column shows the corresponding clustering results after edge removing and root searching. Except the last dataset (bottom), the Edge Plots of all the other datasets are salient enough that the undesired edges for these datasets are easy to be determined. Unlike K-means for which the cluster number *K* needs to be specified in advance, here the cluster number (one less than the number of the undesired edges) can be determined interactively. In other words, here the "cake" (referring to the IT structure) is made in advance and how to cut the cake for users is also well indicated by the "instruction" (referring to the Edge Plot). Note that here we used the K-NN ($K = 10$) in the first step of H-NND and the Euclidean distance to measure the distance (or the dissimilarity) of any pair of data points. See the K-NN graphs and some intermediate results in the supplementary material (Fig. S1) attached to the end of the paper.

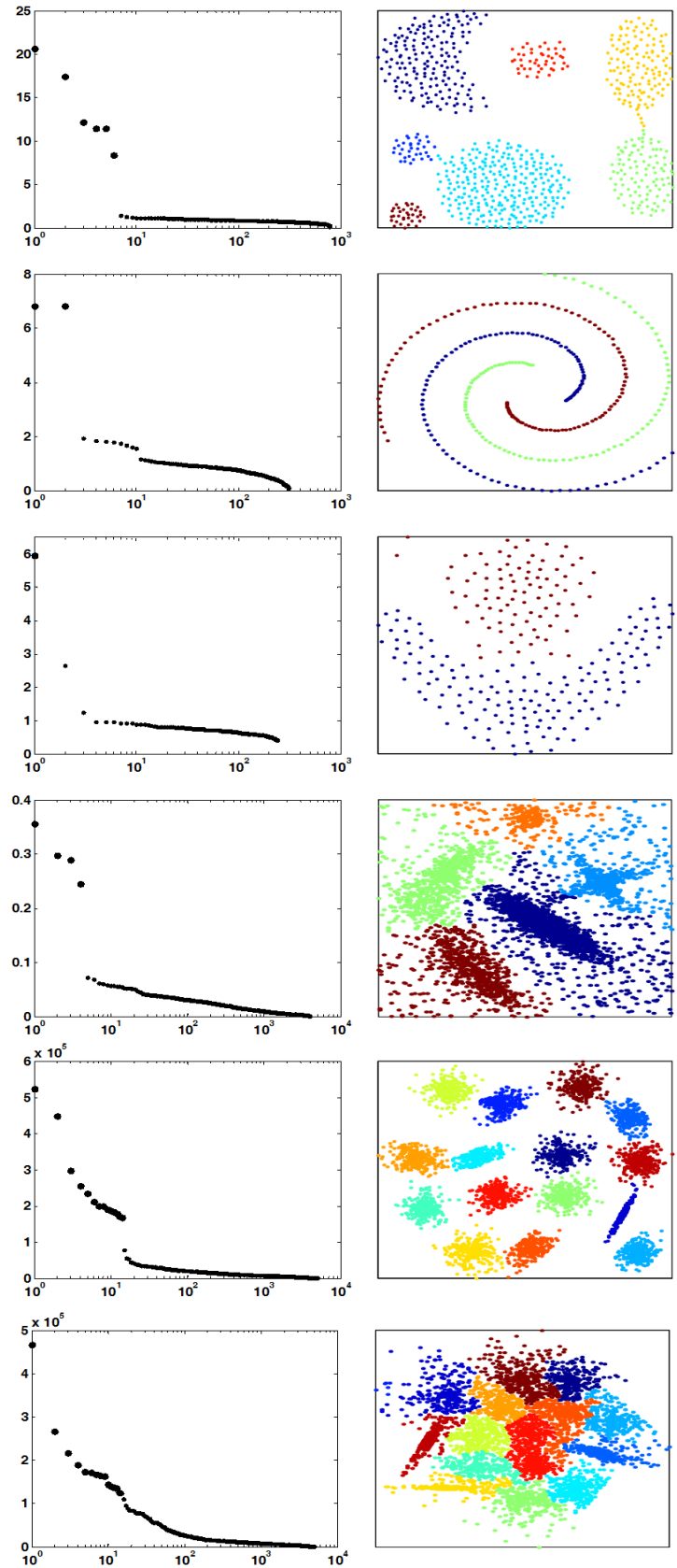

**Fig. 4. Results on six synthetic datasets by H-NND**. **Left column**: the Edge Plots. Horizontal axis denotes the edge length. **Right column**: the corresponding clustering results.

## 4.2 Tests on high-dimensional data sets

In this set of tests on the Iris, D32 and Mushroom datasets, the kernel-based method was used to compute the potentials on nodes, with σ = 0.0001, 1 and 4, respectively. And K-NN was chosen in step 1 to construct the initial neighbor graph. Besides, for Iris and D32, the cosine and Euclidean distance were used to measure the pair-wise distance, respectively. For Mushroom dataset, since the attributes of the data instances are characters rather than real values, we took the number of different characters from any pair of data instances as the distance. The test results are as follows.

For Iris dataset, only 3 out $N = 150$ data instances were falsely assigned into 3 clustered when $K$ was set as any integer from 1 to $N - 1$, except 2. When $K = 2$, this error number was slightly increased to 6. Fig. 5A shows one typical Edge Plot of this dataset.

For D32 dataset, almost any $K$ can achieve the ideal result, with all data points being correctly assigned into 16 clusters. See Fig. 5B for one Edge Plot of it.

For Mushroom dataset, the performance was also comfortable. When $K = 5$, the cluster error rate was 2.76% with 25 clusters. However, when $K = N - 1$ (in this case, the neighborhood constraint loses its meaning and thus H-NND, NND and ND become equivalent), the performance became better: 23 clusters with zero error rate. Therefore, although H-NND is designed to make the undesired edges more salient than that of ND, we cannot say H-NND can be always superior to ND. The reason behind it needs further consideration.

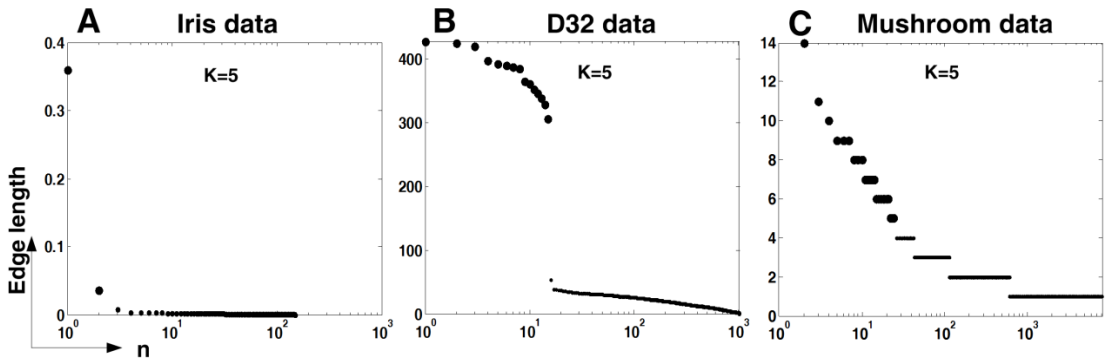

Fig. 5. The Edge Plots of three high-dimensional datasets.

## 4.3 Experiments to compare MST, ND and H-NND

In Fig. 6, we made comparisons between the minimal spanning tree (MST) algorithm and our algorithms ND and H-NND, since, for clustering purpose, the graphs generated by them (MST, IT and IT, respectively) all involve the redundant edges that requires to be removed.

Compared with the IT structure constructed by ND (Fig. 6B), the redundant edges (in red) in the IT structure constructed by H-NND in Fig. 6C are overall more salient, some being much longer at least. This is also demonstrated in the corresponding Edge Plots in Fig. 6 E and F, in which we can see that the average gap between the points with the largest edge lengths with the others in Fig. 6F is obviously larger than that in

Fig. 6E. However, the overall performances for both K-NND and ND are much better, as compared with those (Fig. 6A) in the MST. The Edge Plot (Fig. 6D) for the MST also shows smaller average gap between the points with the largest edge lengths with the others. It is worth noting that there are only 4 points (rather than 6) popping out in Fig .4D, which is in line with the fact in Fig. 6A that two undesired edges are of very short lengths, even shorter than most non-redundant edges. That is why, for the MST, even when we chose to cut off the 6 longest edges, this still leads to a poor clustering result as shown in Fig. 6G, where there are two fake clusters (denoted by the red circles) and two clusters (in green and blue) not segmented well. In contrast, for our ND and H-NND methods, the number of the undesired edges or the threshold to cut the edges is easy to be determined as Fig. 6 E and F show, and the edge-removing process proves very accurate, as the sound clustering results in Fig. 6 H and I present.

In Fig. 6, we used DT in the first step of H-NND. In fact, comparable results were still obtained when K-NN was used (See supplementary material Fig. S2).

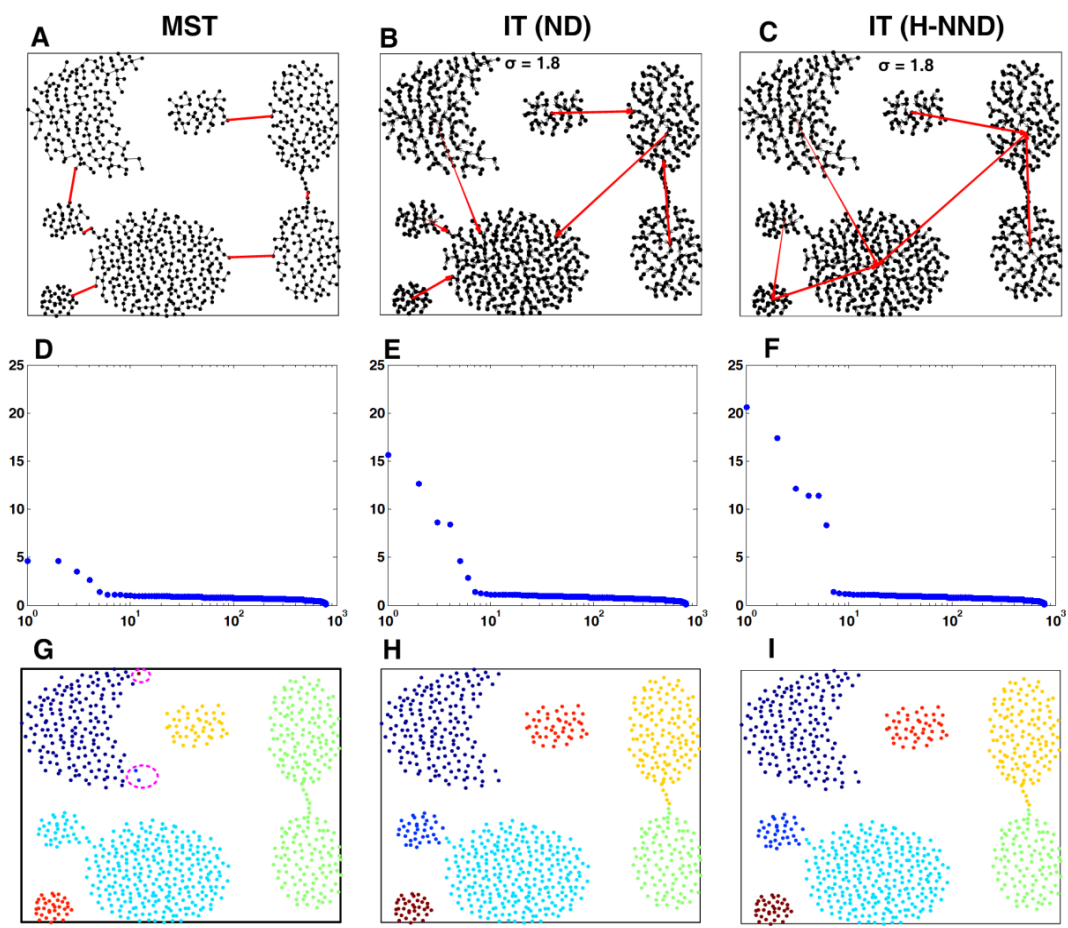

**Fig. 6. Comparison between the MST, ND and H-NND methods**. Columns from left to right correspond to the results of the MST, ND and H-NND methods, respectively. The comparison is taken from three aspects: the generated graphs (1st row), the Edge Plots (2nd row) and the clustering results after removing the undesired edges (3rd row). The potentials here for ND and H-NND were both computed by the kernel-based method with the same parameter σ = 1.8.

### 4.4 Experiments to compare NND and H-NND

In Fig. 7, we made some experiments to further demonstrate the fact that H-NND can effectively solve the over-partitioning problem of NND. The experiments are mainly focused on the 6th dataset in Table 1. Although NND could achieve the same performance with H-NND when $\sigma = 10000$ and $K = 10$, when $\sigma$ was reduced to 5000, or K was decreased to 5, the over-partitioning problem of NND emerged, with respectively $C = 57$ (Fig. 7 Ai) and $C = 64$ (Fig. 7 Bi) clusters, far more than the true cluster number ($C = 15$). In comparison, almost invariably ideal results were achieved by H-NND (Fig. 7 Aii and Bii). Furthermore, when we tried to reduce the number of parameters, such as, (i) eliminating the parameter $\sigma$ by using the local information in the neighborhood graph as we introduced in (14) to estimate the potentials on nodes, and (ii) further eliminating the parameter K by using the DT (nonparametric) rather than K-NN as the initial neighborhood graph, the over-partitioning problem became worse for NND, as shown in Fig. 7 Ci and Di with as many as $C = 258$ and 899 respectively. This is not surprising at all, since the above non-parametric approaches will make the estimated distribution of the potentials less smooth, and thus more locally extreme nodes will be generated and taken as cluster centers. However, this still made almost no impact on the results (Fig. 7 Cii and Dii) of H-NND. Moreover, the Edge Plots for H-NND (Fig. 7 Aiii, Biii, Ciii and Diii) are all quite salient and thus the undesired edges and their numbers can be easily and reliably determined.

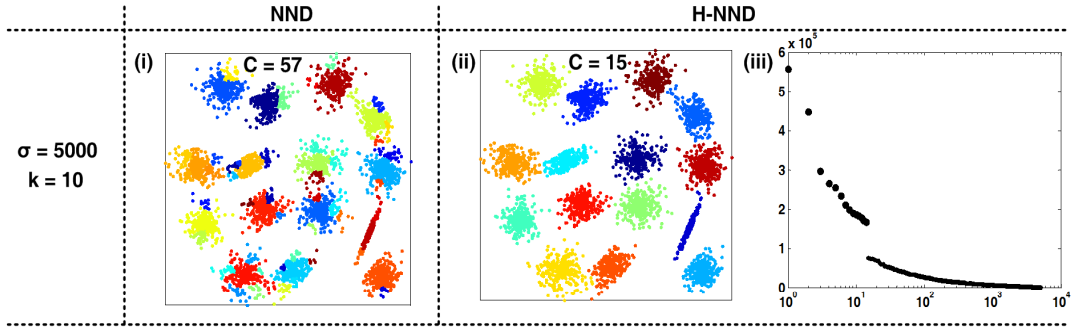
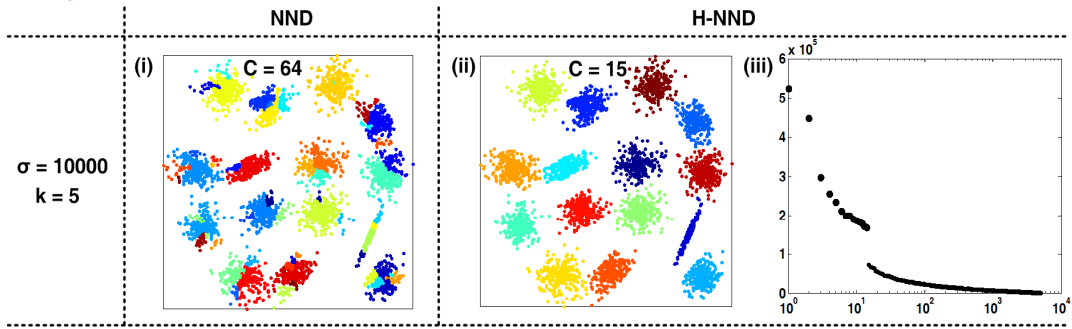
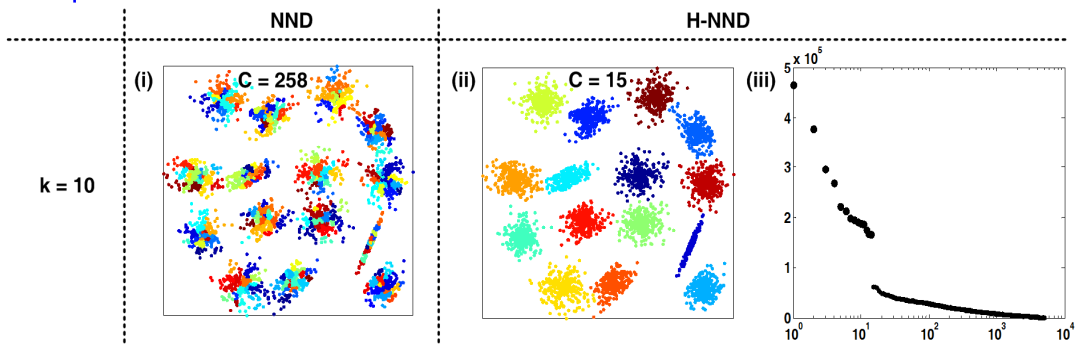
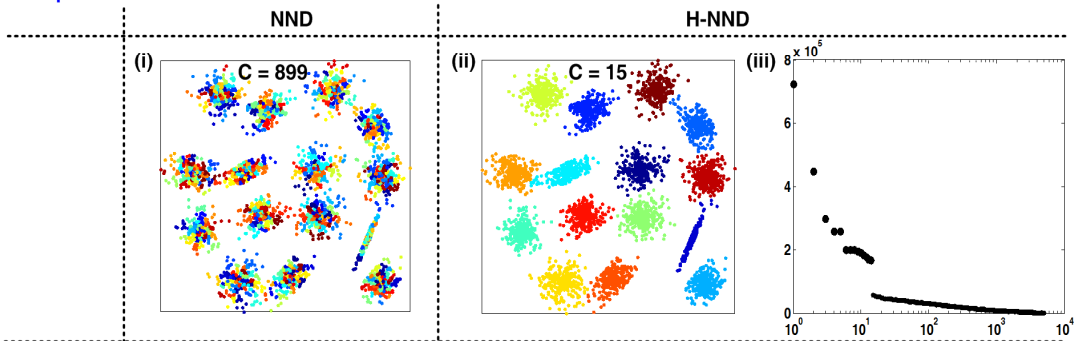

*C: cluster number

**Fig. 7 NND vs. H-NND**. As the number of parameters involved in the methods is gradually decreased to zero in (D), the over-partitioning problem for NND gets worse, reaching a large number of *C* = 899 clusters in (D), while the performance for H-NND remains constant at *C* = 15 clusters.

## 5 Conclusions and Discussions

**5.1 H-NND vs. ND**

H-NND has some advantages over ND in terms of the computation time and the saliency of the redundant edges:

(i) Compared with ND**, *H-NND significantly reduces the computation time of determining the parent nodes***. This can be explained from two aspects. ***For the non-extreme nodes*** (in the first layer), due to the neighborhood constraint, only the local computations are involved in determining their parent nodes (i.e., step 3). This is in stark contrast with ND, which needs to search the whole dataset to determine the parent nodes of those nodes. ***For the extreme nodes*** (in the second layer), although the process (i.e., step 4) of searching their parent nodes follows the process of ND without the neighborhood constraint, the search scopes are just among these extreme nodes themselves (a very small sub-dataset) rather than the whole dataset again. For instance, H-NND took 0.1s in step 3 and 0.0004s in step 4 when testing the Mushroom dataset ($N = 8124$) in a personal computer (Intel Core i5, 3GHz with 8GB RAM). Namely, given the neighborhood graph, it took 0.1004s for H-NND to construct the IT data structure for this dataset, much faster than ND (given the pair-wise similarities, ND took ~4.5s to construct the IT data structure). Besides, note that after removing the redundant edges, the computation time for the remaining process of searching the root nodes (or clustering assignment process) is generally negligible [e.g., no more than 0.005s on this dataset, see Table S2 in (1)]. Thus, given the neighborhood graph, it costs only around 0.1s for H-NND to output the clustering results (ignoring the interactive process of determining the redundant edges).

(ii) Compared with ND, ***H-NND significantly increases the saliency of the redundant edges while making no or insignificant change on the non-redundant edges*** (see Fig. 6 and Fig. 3). The reasons for this are analyzed in this way. First, the datasets can be divided into three classes: the non-extreme nodes, the non-valid extreme nodes and the valid extreme-nodes. And accordingly, there are three classes of edges, i.e., edges started from the non-extreme nodes, the non-valid extreme nodes and the valid extreme-nodes, respectively. The first two classes of edges are usually the non-redundant edges, and the third class of edges are the redundant edges. Compared with the IT generated directly by ND, the first class of edges have almost no change (in terms of their lengths), since the parent nodes (end nodes) of the non-extreme nodes (start nodes) will hardly be affected by the neighborhood constraint. For the second class of edges, there should exist small changes, yet usually insignificant, if the density is roughly captured (Fig. 3). For the third class of edges, i.e., the redundant edges, both their start and the edge nodes are the extreme nodes, in contract to ND for which the end nodes of the generated redundant edges are usually not the extreme nodes. Since the extreme nodes for these redundant edges are usually in the center of the clusters, this can make the redundant edges more distinguishable (Fig. 6). In conclusion, compared with ND, H-NND makes no or insignificant changes to the non-redundant edges while making significant changes to the redundant edges, and thus the redundant edges become more distinguishable. One risk behind H-NND is that the second class of edges could be comparable to the third class of edges if the density is badly estimated.

### 5.2 H-NND vs. NND

Compared with NND, clustering by H-NND largely resolves the over-partitioning problem of NND (Fig. 7). NND provides a simple and effective way that is able to fast and simultaneously find all local extreme points in the estimated potential function. While NND is faced with the over-partitioning problem, H-NND is able to further distinguish from all the extreme points detected by NND the valid ones to be the ultimate cluster centers. This is in fact the common advantage shared by ND. Note that, like ND and NND, H-NND not only refers to the process of constructing the IT, but also the short term of the clustering by H-NND.

### 5.3 The view from ND to NND and H-NND

Interestingly, the development from ND to NND then to H-NND here is quite consistent with the law of "*the negation of the negation*" (denoted as ***Double Negations***, see Text S1 in the supplementary material) formulated by the philosopher *G. Hegel*. In NND, we realized the "mistake" we made in ND, i.e., not taking the constraint of the physical space into account, and thus we added this neighbor graph constraint (approximating the physical space). Consequently, the redundancy (referring to the redundant edges) was prevented. *This could be viewed as **the 1st Negation.*** It refers to the negation of not choosing the constraint, or the negation of choosing the redundancy in ND. However, NND didn't make things better. Therefore, in this paper (H-NND), we abandon the constraint and come back to ND again, while this time ND is only performed on a subset of the original dataset, i.e., the root nodes resulted from NND. Like ND, H-NND also generates the IT structure, in which the redundant edges once again appear. *This could be viewed as **the 2nd Negation.*** It refers to the negation of choosing the constraint or the negation of abandoning the redundancy in NND. After the 2nd negation, things become better. Compared with ND, the redundant edges in the IT structure generated by H-NND generally become more salient, thus being much easier and more reliable to be identified. In other words, the IT structure constructed by H-NND becomes more fitted for data clustering.  \

### 5.4 The view from H-NND to NND and ND

Also interestingly, if we look the above development inversely, namely from H-NND to NND and then to ND, we can see that, some "constraints" require to be broken so as to reach the simple yet powerful ND rule. In other words, H-NND indirectly demonstrates the simplicity and effectiveness of ND or Rodriguez and Laio's *Decision Graph* (DG) (15) (ND can be viewed as another form of DG).

In our perspective, comparing with adding constraints, it could be more challenging yet meaningful to find out the "constraint" or the fixed thinking modes in our mind and break it. For instance, although we have already done several works in this theme, we surprisingly find till now that we made a mistake unconsciously in our latest paper (4) (see its first preprint version in arXiv.org), where we redundantly added the term "*neighbor*" while introducing ND and consequently ND became NND there. Another example is about the "*Gradient*". Together with the associated methods "Gradient Ascent" or "Gradient Descent", gradient is often used in the Gradient-based clustering methods. Nevertheless, gradient describes the ***local feature***, and thus, just like the ***neighborhood constraint*** in NND, those Gradient-based methods should also face the same over-partitioning problem as NND does[4]. We will discuss more about this in the next paper.

Generally speaking, two elements, from the view of mythology, are quite contributing in H-NND: the ***hierarchical strategy*** and the ***neighborhood constraint***. Under the guidance of hierarchical strategy, H-NND does not attempt to construct the IT structure in one stage, which makes it possible for the role of NND to be fully played. And the effectiveness of the hierarchical strategy is guaranteed by the neighborhood constraint, since the neighborhood constraint affects only the extreme nodes by narrowing the choices of their parent nodes in local scopes and consequently they become the root nodes of the generated sub-graphs and are postponed to be processed in the second stage. From another perspective, we can say that, since the neighborhood constraint can tell the extreme nodes from the whole dataset beforehand, we can thus deal with the extreme and non-extreme nodes separately in different stages. And in fact, as far as we know, the ***neighborhood constraint*** is a very useful strategy that also benefits lots of non-linear dimensionality reduction methods like Isomap (16), LLE (17), TPE (18), etc.

**Acknowledgements**: We thank Prof. Shuicheng Yan for the helpful discussions.

---

[4] Theoretically, we can promote the performance of those Gradient-based methods similarly as what we do in H-NND here. But the performance is not as well as the one showed here, since the Gradient-based methods is sensitive to the parameters and thus cannot provide a reliable basis in each layer. We will show this in the next paper.

**Supplementary Material (*Fig. S1, Fig. S2 and Text S1*)**

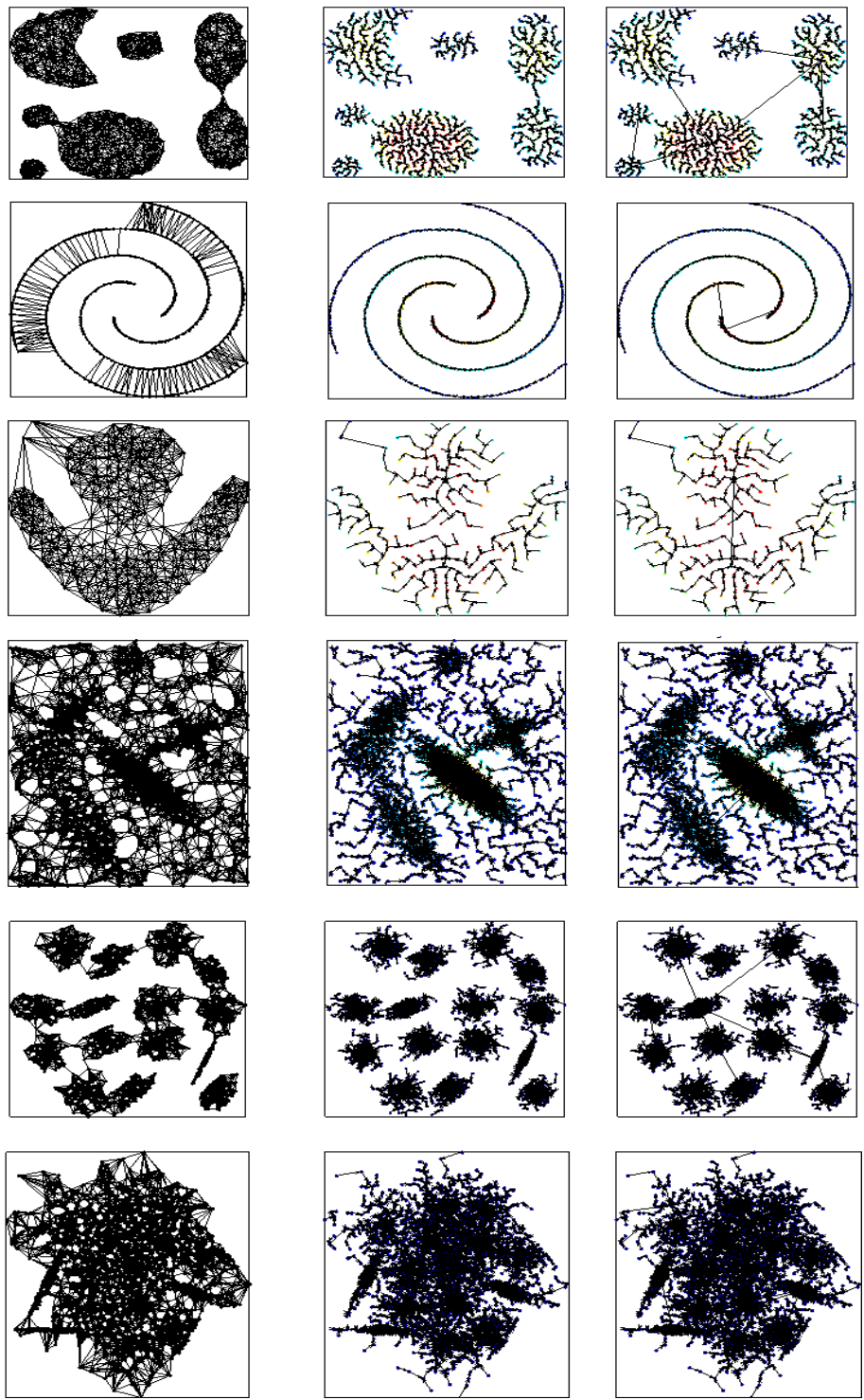

**Fig. S1** First column: K-NN (*K* = 10) graphs. Second column: results of NND. Third column: the IT structures made by H-NND. The kernel-based method is used to compute the potential with the parameter *σ* = 1.8, 1.8, 1.8, 0.005, 10000, 10000.

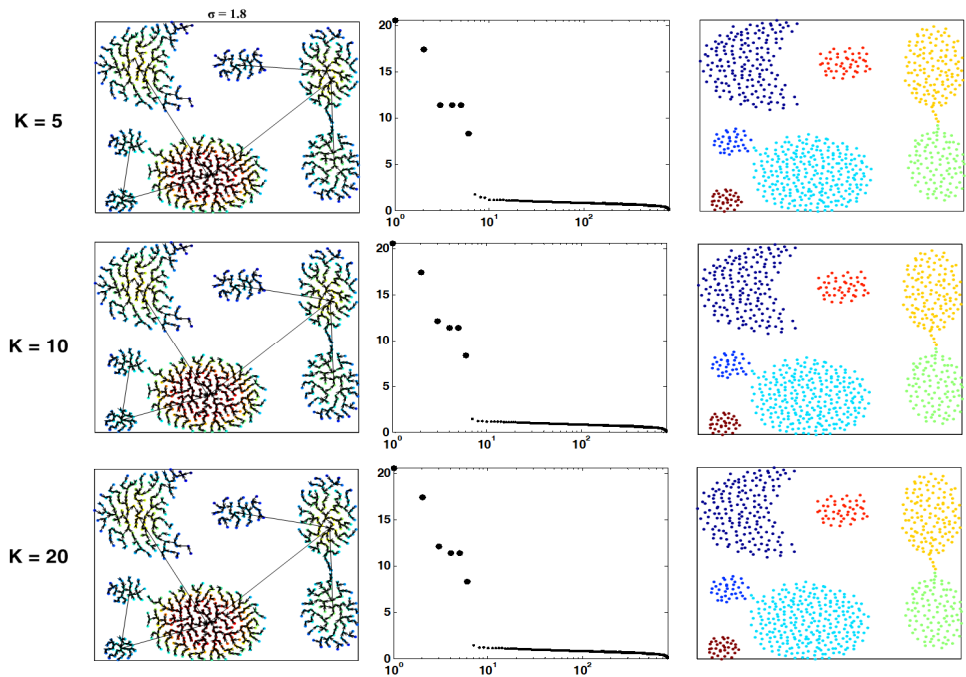

**Fig. S2 The results when K-NN (with varying K) was chosen as the neighborhood graph for H-NND.** Rows from up to bottom correspond to the results with $K = 5, 10, 20$, respectively. In fact, even when $K$ is increased to $N - 1$, H-NND could still perform well, since in this case H-NND is reduced to ND and thus the results are just what the 2nd column of Fig. 6 shows. Note that we used the kernel-based ($\sigma = 1.8$) to compute the potential.

### Text S1 "The Negation of the Negation"

"……The law of the negation of the negation was first formulated by G. Hegel, but particular features of it had previously been established in philosophy (the dialectical character of negation, the role of continuity in development, and the nonlinear character of the direction of development). In Hegel's dialectical system, development is the emergence of a logical contradiction and its subsequent sublation. In this sense, *development is the birth of the internal negation of the previous stage, followed by the negation of this negation* (G. Hegel, Soch., vol. 6, Moscow, 1939, pp. 309–10). *To the extent that the negation of the previous negation proceeds by sublation, it is always, in a certain sense, the restoration of that which was negated, a return to a past stage of development. However, this is not a simple return to the starting point, but "a new concept, a higher, richer concept than the previous one, for it has been enriched by its negation or opposite; it contains in itself the old concept, but it contains more than this concept alone, and it is the unity of this and its opposite"* (ibid., vol. 5, Moscow, 1937, p. 33). Thus, the law of the negation of the negation is the universal form of the splitting of a single whole and the transition of opposites into each other— that is, the universal manifestation of the law of the unity and struggle of opposites……"

*From*: <u>http://encyclopedia2.thefreedictionary.com/Negation+of+the+Negation,+Law+of+the</u>